\documentclass[runningheads]{llncs}
\usepackage{graphicx}
\newif\iffinal
\finaltrue

\usepackage{listings}
\usepackage{hyperref}
\usepackage[capitalise]{cleveref}
\usepackage[T1]{fontenc}

\usepackage{stackengine}
 
\usepackage{xcolor}

\begin{document}
\title{Volkit: A Performance-Portable Computer Vision Library for 3D Volumetric Data}
\titlerunning{Volkit Image Manipulation Library for Volumetric Data}
%
\iffinal
\author{Stefan Zellmann\inst{1}\orcidID{0000-0003-2880-9090} \and
Giovanni Aguirre\inst{2}\orcidID{0000-0001-7999-8528} \and
J\"urgen P. Schulze\inst{2}\orcidID{0000-0003-4903-4837}}
\authorrunning{S. Zellmann et al.}
%
\institute{University of Cologne, Weyertal 121, 50931 Cologne, Germany
\email{zellmann@uni-koeln.de}\\
\and
UC San Diego, 9500 Gilman Drive, La Jolla, CA 92093, USA\\
\email{\{g3aguirre,jschulze\}@ucsd.edu}}
\fi
\maketitle              

\begin{figure*}[h]
\vspace{-2em}
\stackunder[2pt]{\stackunder[2pt]{\includegraphics[width=0.06\textwidth]{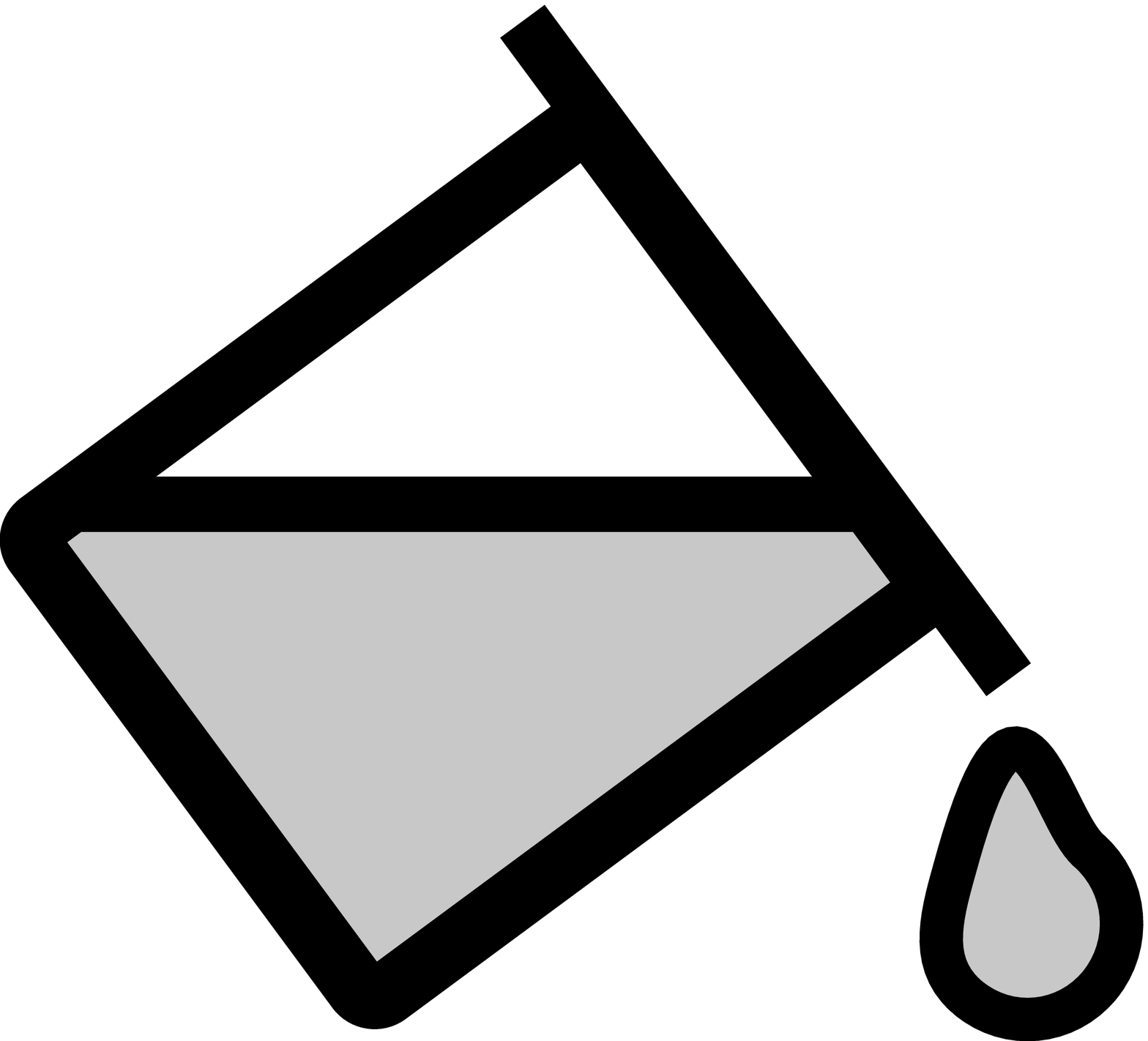}}
{\includegraphics[width=0.245\textwidth]{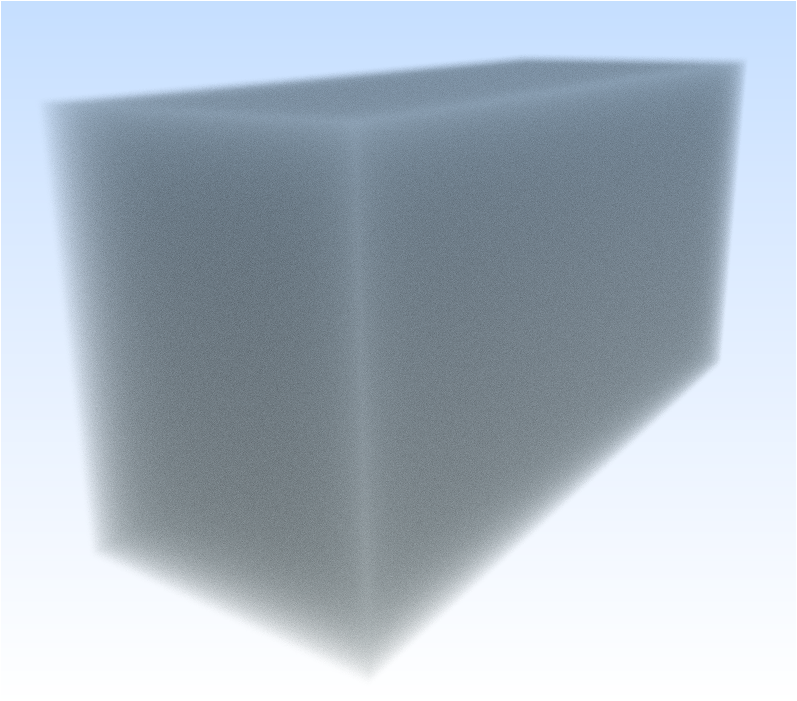}}}{\texttt{Fill()}}
\stackunder[2pt]{\stackunder[2pt]{\includegraphics[width=0.06\textwidth]{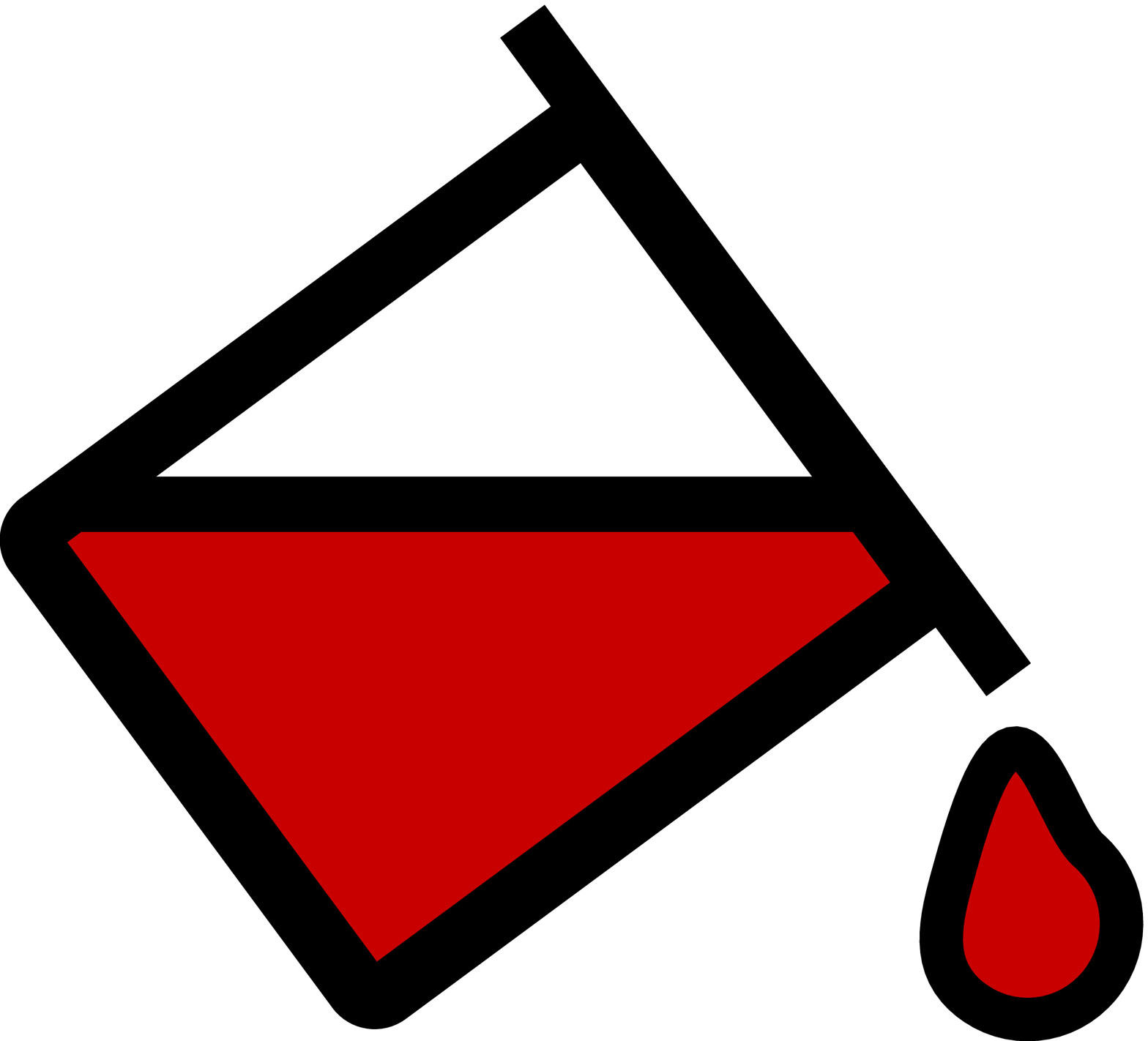}}
{\includegraphics[width=0.245\textwidth]{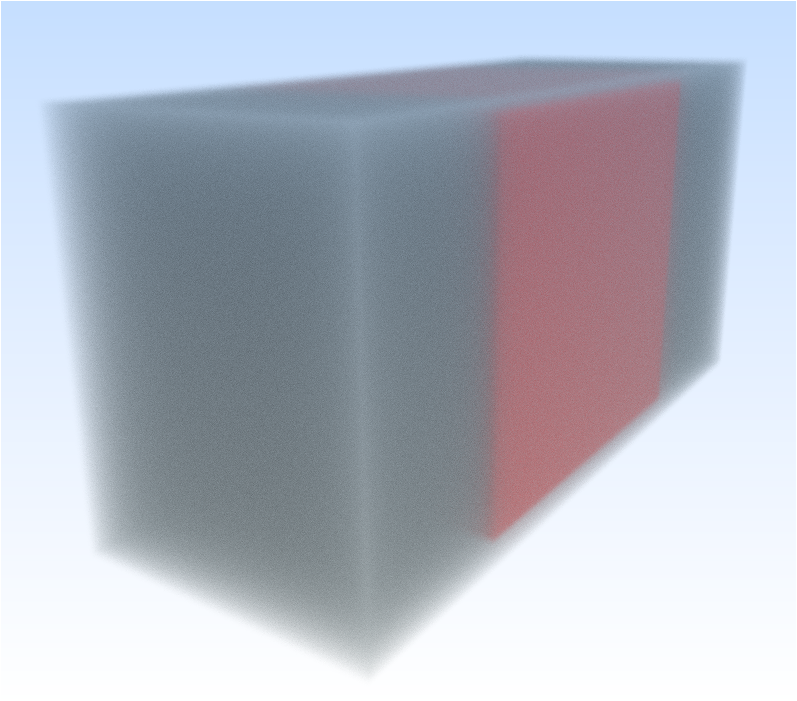}}}{\texttt{FillRange()}}
\stackunder[2pt]{\stackunder[2pt]{\includegraphics[width=0.05\textwidth]{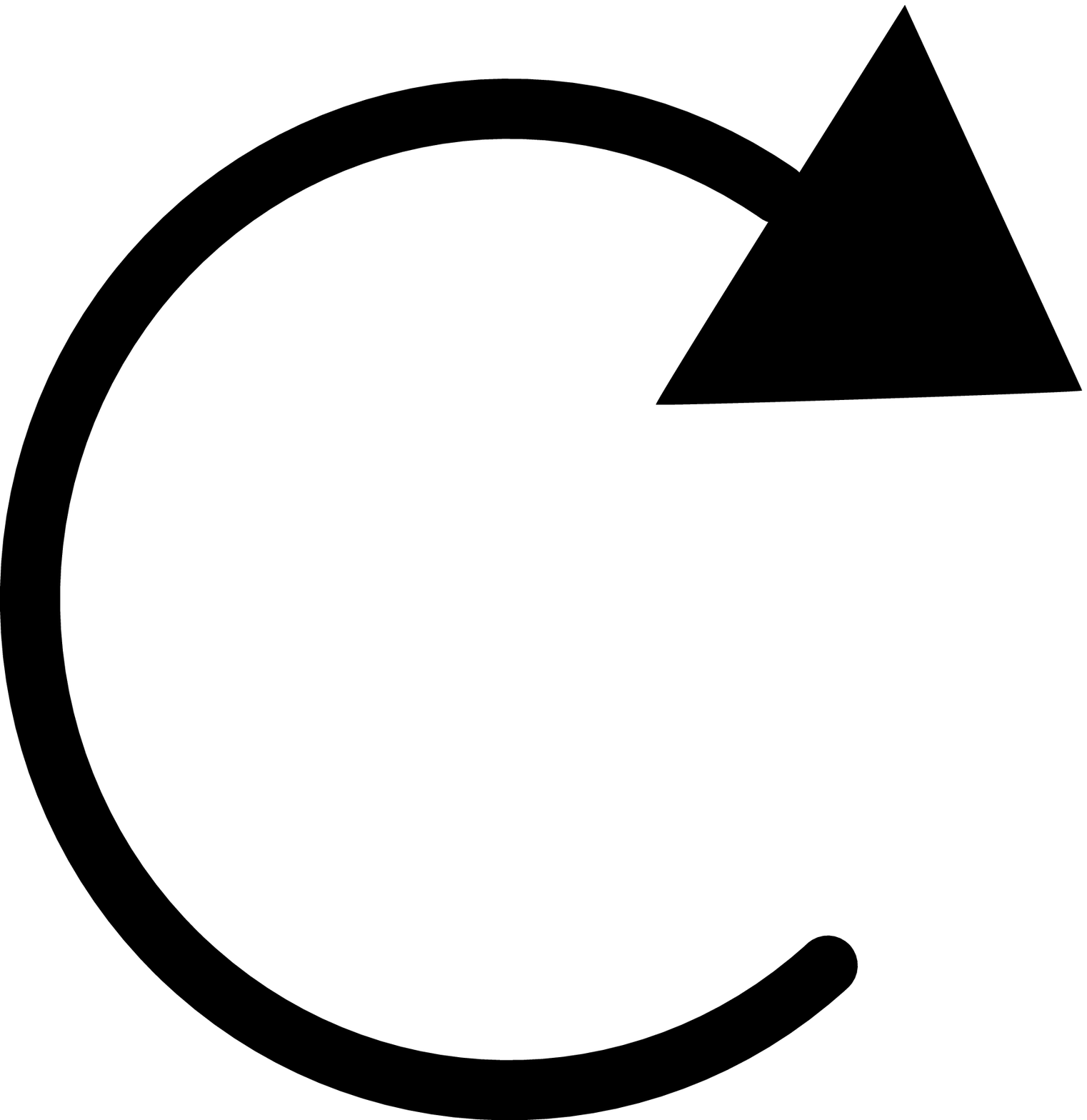}}
{\includegraphics[width=0.245\textwidth]{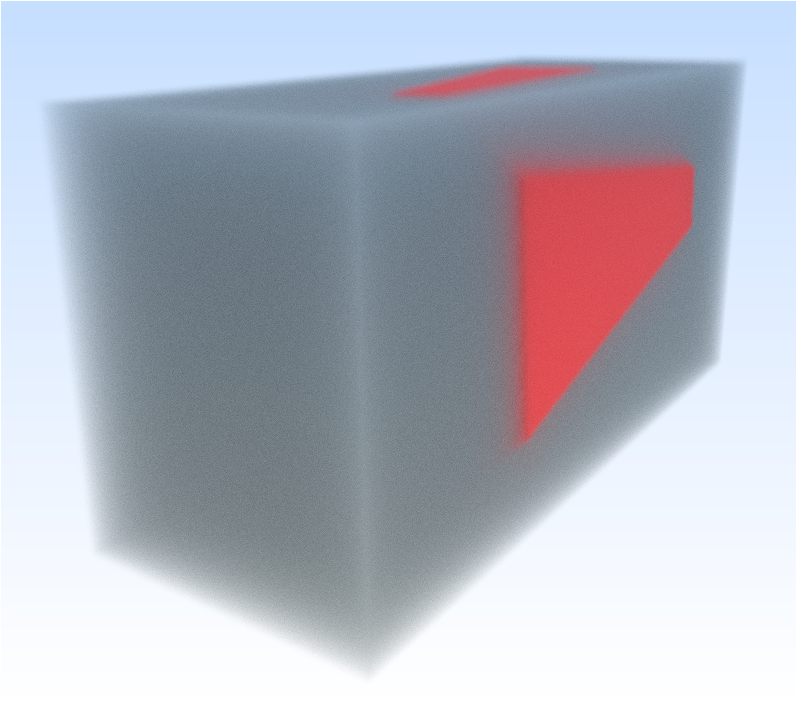}}}{\texttt{RotateRange()}}
\stackunder[2pt]{\stackunder[2pt]{\includegraphics[width=0.06\textwidth]{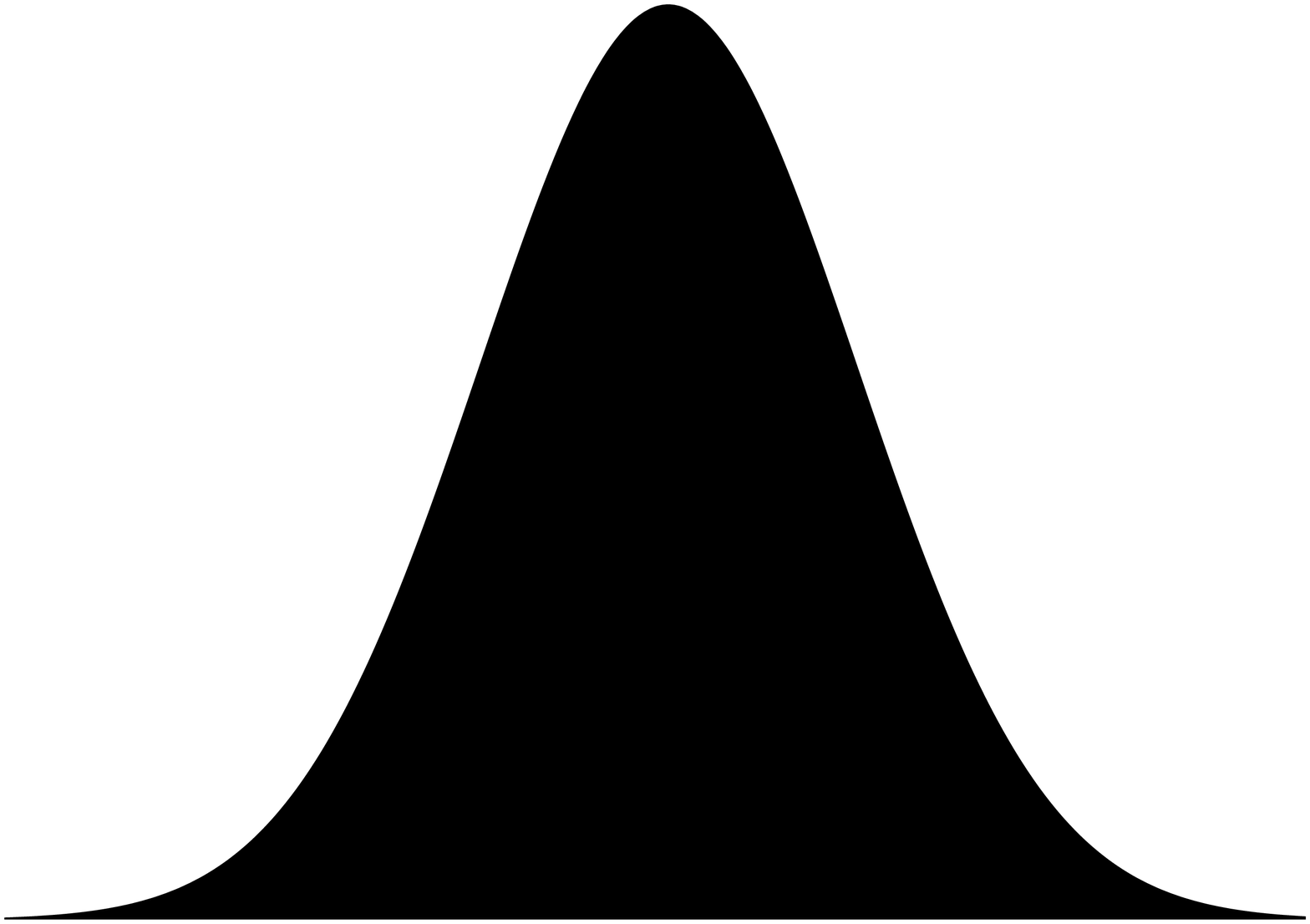}}
{\includegraphics[width=0.245\textwidth]{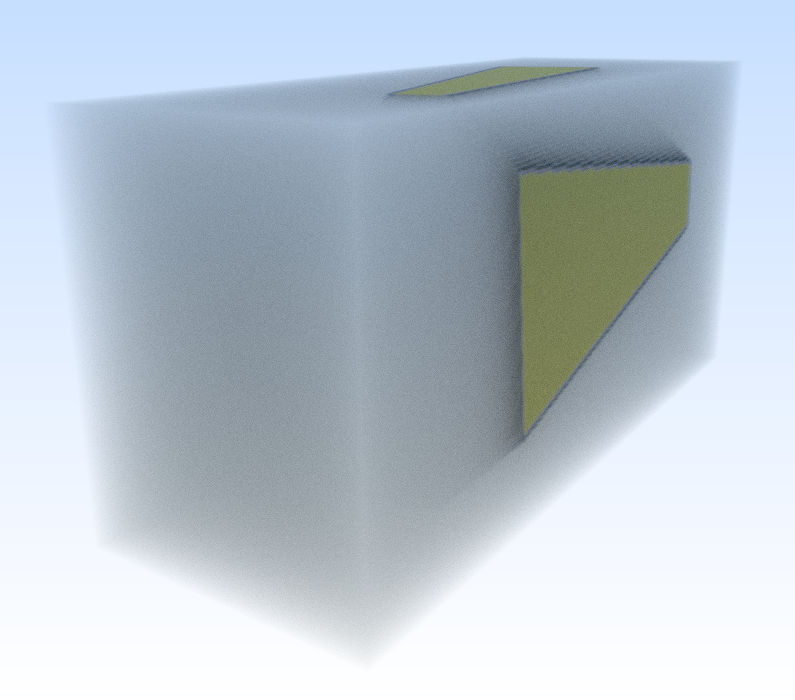}}}{\texttt{ApplyFilter()}}
\caption{\label{fig:teaser}
Exemplary image manipulation session with Volkit. We use the 
\texttt{Fill} algorithm to fill the 3D volume with gray, then use the
\texttt{FillRange} and \texttt{RotateRange} algorithms to fill and rotate the red
region with subvoxel accuracy, and finally apply Gaussian blur with
\texttt{ApplyFilter}. Compared to other libraries, our library focuses
on image manipulation routines for 3D volumes and is cross-platform and 
performance-portable.}
\vspace{-3em}
\end{figure*}

\begin{abstract}
We present volkit, an open source library with high performance implementations
of image manipulation and computer vision algorithms that focus on 3D volumetric representations. Volkit
implements a cross-platform, performance-portable API targeting both CPUs and GPUs
that defers data and resource movement and hides them from the application developer using a 
managed API. We use volkit to process medical and simulation data that is 
rendered in VR and
consequently integrated the library into the C++ virtual reality software CalVR.
The paper presents case studies and performance results and by that demonstrates the
library's effectiveness and the efficiency of this approach.


\end{abstract}
\section{Introduction}
Volumetric data representations are prevalent in the simulating sciences as well as in the
clinical daily routine. While design studios or 3D artists often use dedicated
graphics workstations, typical clinical IT systems have restricted hardware
capabilities, and simulation codes are run on supercomputers that use compute
GPUs, or CPUs with instruction set architectures like PowerPC or ARM that
are best targeted with vendor-specific APIs. 

Computer vision libraries like OpenCV~\cite{bradski:2013} or the Insight Toolkit~\cite{johnson:2015}
focus on 2D images, are optimized for single platforms, and don't support hierarchical or unstructured volume
data. As such there is still a lack of a 
general, cross-platform, \emph{performance-portable} library for 3D image processing and computer 
vision tasks (cf.~\cref{fig:teaser}) using modern concepts such as the ones described below.

The simulating sciences often employ a post processing pipeline at the of the
exploration process that does or does not involve 2D or 3D visualization,
but that usually requires data wrangling to transition the data from one pipeline---the data
source---to the other one that is concerned with post processing.
It is this interface between data acquisition and post processing
where we position our cross-platform volume manipulation library 
\emph{volkit}.

Volkit was designed to accommodate the typical data wrangling, data filtering and post processing
phases that most pipelines involve \emph{prior to} data presentation and as such
implements tasks that most typical interactive 3D or virtual reality rendering systems would implement.
Volkit aims at both high performance, high bandwidth data processing
\emph{and} at versatility and ease of use.

Volkit is not a 3D volume rendering library in particular---although it comes with a
high quality path tracing renderer that \emph{might be} suitable for
rendering in many cases---yet we acknowledge that volumetric image processing
\emph{eventually} may culminate in the user visually exploring their data sets.
Visual exploration, and in particular interactive 3D visualization \emph{is} a part of
a typical scientific post processing pipeline, and we therefore also integrate and
evaluate volkit in the context of the virtual reality and scientific visualization
software CalVR~\cite{schulze:2013}.


\section{Related Work}
The most common volumetric data representations nowadays are structured and unstructured~\cite{schroeder:2005}
as well as hierarchical grids---e.g., Octrees~\cite{burstedde:2011} and other, often
regular trees with wider or variable branching factors~\cite{fryxell:2000}.
While structured grids are most
frequently generated by microscopes and CT or MRI scanners, unstructured and hierarchical grids---the
latter are also referred to as \emph{adaptive mesh refinement}
(AMR)~\cite{berger:1984,berger:1989}---are often the result of simulation codes.
Typical codes that output AMR data are for example
FLASH~\cite{fryxell:2000} or \texttt{p4est}~\cite{burstedde:2011}.

The image processing library OpenCV~\cite{bradski:2013} provides
many filtering and contrast enhancement algorithms, but
doesn't support volumes.
ITK~\cite{johnson:2015} \emph{does}
support images of higher dimension than 2D, but
treats those as if
they were 2D images and hence has no notion of the topologies described above that are
arguably more common in 3D than they are in 2D. Furthermore, ITK is targeted
at CPUs only. The volume manipulation tool 3D~Slicer~\cite{kikinis:2014}
has a Python API, but is not
performance-portable and only supports structured volumes.

While performance-portable volume manipulation libraries for arbitrary to\-po\-lo\-gies
are scarce, there are other GPU libraries hat volkit draws
inspiration from. The deferred API for example was inspired by the OptiX ray tracing
library~\cite{parker:2010}. OptiX's \emph{node graph} 3D scene abstraction
uses group nodes with
\emph{accelerators} (bounding volume hierarchies) and optional data buffers.
The user specifies \emph{which} data
is copied to those buffers from the CPU, but not \emph{when} that
happens; similarly, \emph{when} OptiX will (re-)build the accelerator isn't
controlled by the user. OptiX however recently switched to an
explicit API where the user has more control at the cost of increased complexity.
The OptiX 7 Wrappers Library (OWL)~\cite{wald:2020} fills that gap by providing a
node graph API on top of OptiX~7.

Interactive volume rendering is nowadays implemented using ray casting~\cite{levoy:1988}
with absorption and emission~\cite{max:1995}. Optimizations such as early ray
termination~\cite{hadwiger:2006} or empty space skipping
\cite{hadwiger:2018,zellmann:2018,zellmann:2019b} using
real-time ray tracing
hardware~\cite{wald:2021} render this effective even for large
data sets. The
community is however gradually transitioning to volumetric path tracing
\cite{kroes:2012,openvkl:2021}. Volkit's
interactive path tracer was inspired by the
one proposed by Raab~\cite{raab:2019}.
Rendering unstructured grids
requires acceleration data structures~\cite{morrical:2020,wald:2021b}
or connectivity information~\cite{muigg:2011}. Cell-centric AMR volumes require complex 
interpolants for high-quality reconstruction~\cite{wang:2018}.
It has still been shown that in both cases, with carefully designed 
data structures, high-quality interactive rendering is possible~\cite{sahistan:2021,wald:2020b}.


\section{System Overview}
\label{sec:system}
\begin{figure}[tb]
\includegraphics[width=\textwidth]{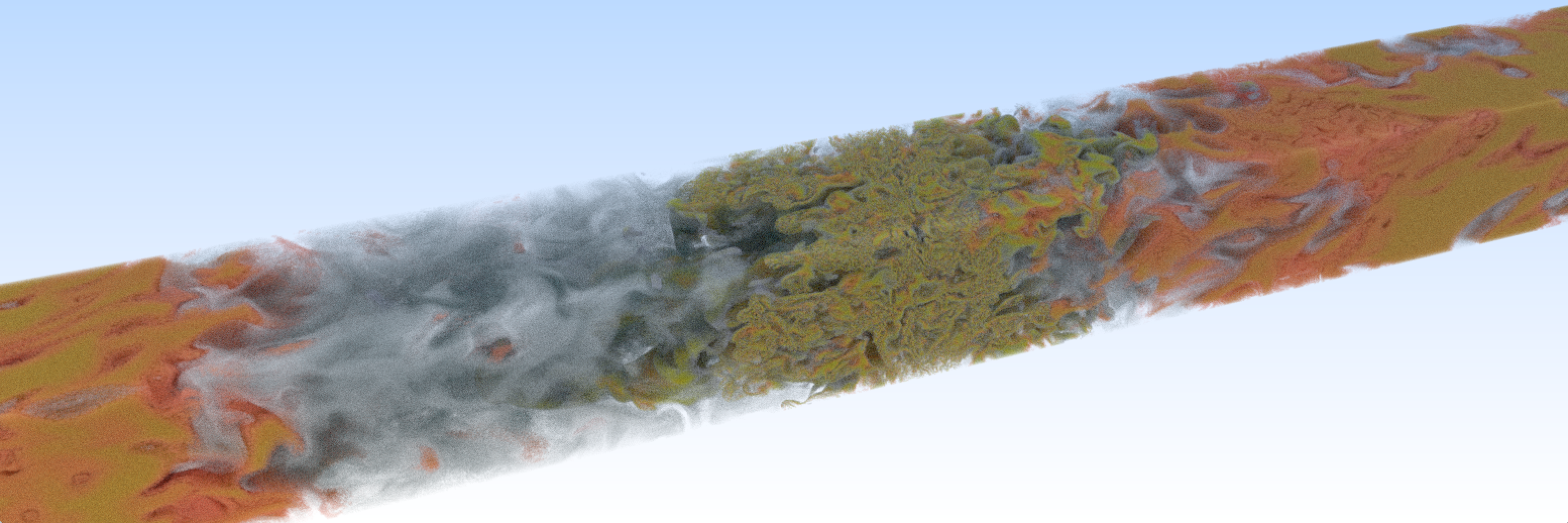}
\vspace{-2em}
\caption{
Path tracing an astrophysics data set using the
\texttt{Render} algorithm. Volkit can render both structured and AMR volumes
using various integrators. Cell-centric AMR volumes are rendered with
high-quality linear interpolation even at level boundaries.}
\label{fig:silcc}
\end{figure}
With data-intensive GPU and coprocessor libraries, developers
have to make certain design decisions that are
centered around data management paradigms (e.g., object-oriented or
data flow-centric), around when data is copied to and from
the device (immediate vs. deferred APIs), and about how much control is given to
the user regarding data handling and control flow. With volkit we decided to
adopt an \emph{implicit}, retained mode API using a data management
layer for volumes and for auxiliary data such as lookup tables.
Volkit uses a data flow-centric paradigm and
provides efficient routines to handle very large data sets.
On the I/O side this is accomplished using streams,
which allow for partial data loading and storing,
the specifics of which are described below.

\subsection{Deferred and Managed Resource Handling}
Volkit uses a deferred memory managed layer for implicit GPU data transfers.
For that we use
a thread-local \texttt{ExecutionPolicy} object that is accessed
via the functions \texttt{SetThreadExecutionPolicy()} and
\texttt{GetThreadExecutionPolicy()}.
By that we acknowledge that users might themselves
run volkit in a multithreading environment. 
The object stores the volkit state associated with the current thread,
such as for example the device type (\texttt{CPU} or \texttt{GPU}).
The \texttt{ExecutionPolicy} is also
used to set
other options, e.g., to enable debug message, or to print out algorithm execution
times to the console.
When the user changes the device settings of the current
\texttt{ExecutionPolicy}, all subsequent algorithms that are executed in
the current thread will run on the device that is currently being set.

Algorithms like \texttt{Render} will
implicitly call the member function \texttt{migrate} on the managed objects.
\texttt{migrate} compares the current
\texttt{ExecutionPolicy} object with the one that
was active when the managed object was last accesses. For that, the last
\texttt{ExecutionPolicy} is carried along by the managed types. If the device
has changed, before the data is accessed, a data transfer is initiated
and the last \texttt{ExecutionPolicy} is updated.
That way, the data is transferred right before it is needed. The user can however
also call \texttt{migrate} themselves and by that control when the actual transfer 
happens. If the current and last \texttt{ExecutionPolicy} are the same,
\texttt{migrate} is a no-op.
\begin{figure}[tb]
    \centering
    \begin{minipage}{0.5\textwidth}
    \begin{lstlisting}[language=c++]
// RGBA tuples 
float rgba[] = { ...  };
// Managed LUT
LookupTable lut(5,1,1,
  ColorFormat::RGBA32F);
lut.setData((uint8_t*)rgba); 
// Initiate rendering
RenderState renderState;
renderState.renderAlgo
  = RenderAlgo::MultiScattering;
renderState.rgbaLookupTable
  = lut.getResourceHandle();
Render(volume, renderState);
    \end{lstlisting}
    \end{minipage}
    \begin{minipage}{0.49\textwidth}
    \includegraphics[width=0.99\textwidth]{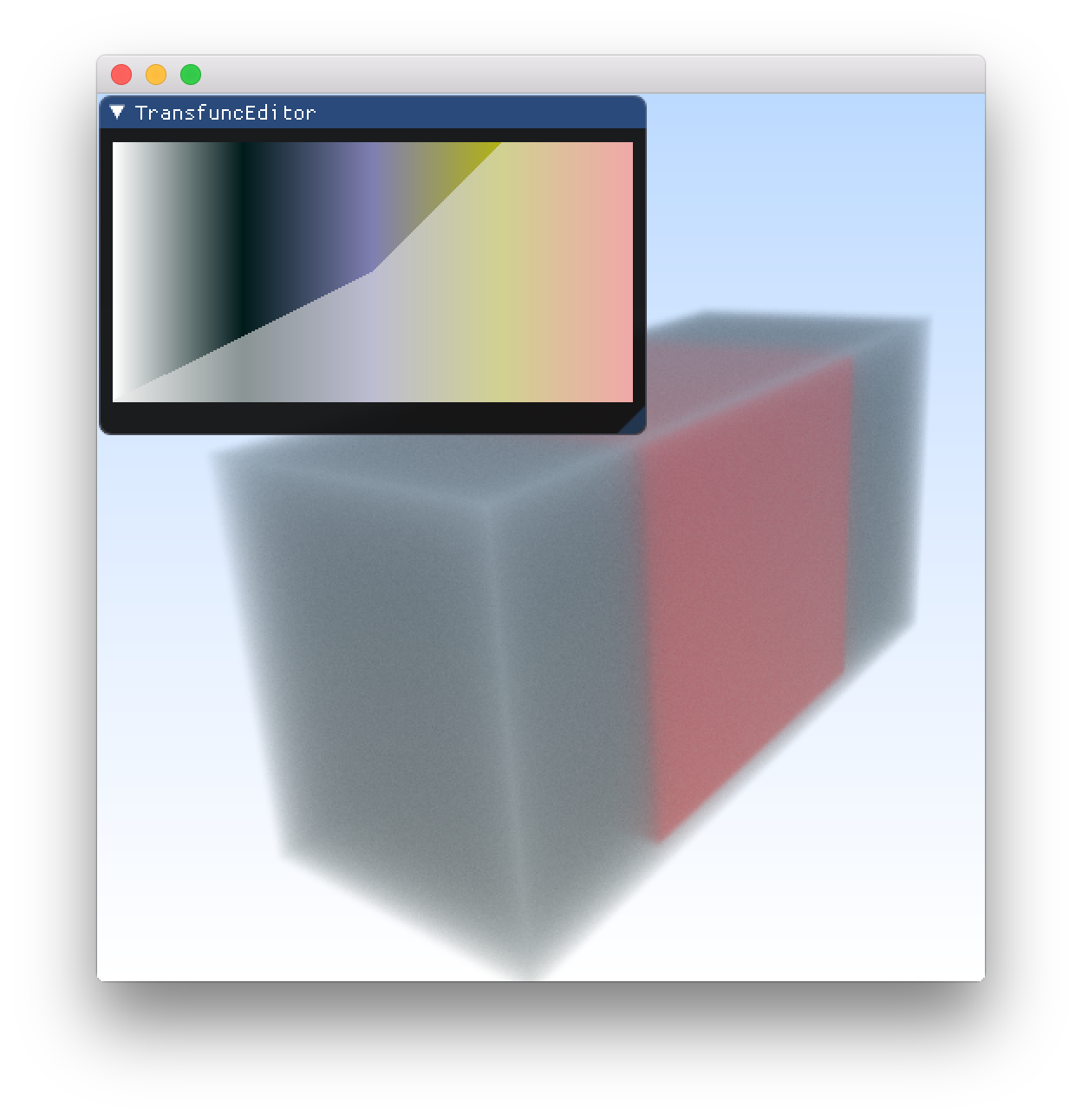}
    \end{minipage}
    \vspace{-2em}
    \caption{
        Interactive ray tracer set up with RGBA lookup table (LUT). The
        user-supplied LUT is used as a transfer function to classify
        samples. LUTs are managed and---as with volumes---are identified
        via their resource handle. The viewer allows
        for rudimentary mouse navigation. The volumetric path tracing renderer
        shown here will interactively render convergence frames.
        \vspace{-1em}}
    \label{fig:render}
\end{figure}
The base class for managed types is called \texttt{ManagedBuffer<T>}
and is templated on the storage type (e.g., \texttt{char} for bytes).
The volume types and other auxiliary buffers
inherit from that template class. Volkit
uses resource handles (cf.~\cref{fig:render}), which are integral types
and allow to identify the managed objects transparently on both
CPU and GPU.



\subsection{Algorithms and Data Structures}
\label{sec:algos}
In the following, we give an overview of the most important data types and
assorted algorithms.\footnote{For a complete documentation see volkit's website
under \url{https://docs.volkit.org}.}
Most of the algorithms that are described below come in two versions: a default
version that operates on the whole volume, and a \emph{range} version
where the identifier \texttt{Range} is appended to the function name. The range
versions operate on a region of interest (ROI) specified via an axis-aligned bounding
box. Furthermore, volkit currently has two
backends for execution on the CPU or on NVIDIA GPUs using CUDA. The
thread-local \texttt{ExecutionPolicy} object from above is used to seamlessly
switch between CPU and GPU execution.

\subsubsection{Volume Types}
Volkit currently supports structured and AMR volumes.
Integrating unstructured volumes would
however be straightforward.
The classes \texttt{StructuredVolume}
and \texttt{HierarchicalVolume} encapsulate the two volume types.
Volkit uses the \emph{basis interpolation method}~\cite{wald:2020b} for
high-quality reconstruction of cell-centric AMR data.
With range algorithms, if the volume is structured,
the ROI is specified in cell coordinates. With AMR volumes,
ROIs are relative to what we call
the \emph{logical grid}---a \emph{hypothetical} structured grid whose size would
allow us to resample the AMR volume without loss of detail.

As proposed by Wald et al.~\cite{wald:2020b}, when loading an 
AMR volume, we drop the AMR hierarchy, keep only the
subgrids, and build a bounding volume hierarchy (BVH) over the \emph{active brick 
regions} (the regions where
the interpolation domains of the subgrids overlap). When accessing the volume, we
use the BVH to perform point sampling using basis interpolation.
We currently support AMR volumes generated by the FLASH simulation 
code~\cite{fryxell:2000}, but integrating other file formats would be 
straightforward. Path tracing of an AMR data set is shown in \cref{fig:silcc}.

\subsubsection{File Handling}
\label{sec:streaming}
I/O is realized using \texttt{InputStream} and \texttt{OutputStream}, which are
passed a \texttt{DataSource} object. Data sources contain headers with meta
information such as volume dimensions, cell sizes,
cell types, or, in the case of AMR volumes the subgrid meta data. Data sources
(which for simplicity also act as sinks)
provide low-level I/O facilities via the functions \texttt{read},  
\texttt{write}, \texttt{seek}, and \texttt{flush} that operate on byte arrays.
In contrast to that, stream objects are aware
of the actual object that is being read or written---their interface is comprised
of functions \texttt{\{read|write\}} whose parameters are \emph{volumes}; the 
functions \texttt{\{read|write\}Range} allow us to support large volumes via partial
loads.

\subsubsection{Volume Manipulation}
\label{sec:imagemanip}
Volume manipulation algorithms are categorized into
\emph{core algorithms} such as, e.g.,
\texttt{Fill}, \texttt{Crop}, \texttt{Resample}, or \texttt{Transform}.
\texttt{Resample} can be used to change the size, the data type,
or the topology of the volume (e.g., to resample an AMR volume on a
structured grid). \texttt{Transform} iterates over all cells,
invokes a user-supplied callback function, and provides that current cell index
and pointer to the cell itself, allowing for arbitrary volume manipulation.

\emph{Derived algorithms} build off of the core algorithms; examples are
\texttt{ApplyFilter} to apply user-specified convolution kernels
(e.g., edge detection or Gaussian blur), which
can be implemented using \texttt{Transform},
or \texttt{Delete}, which is derived from \texttt{Crop}. The significance
here is that the algorithms \emph{can be} implemented using core
algorithms---that doesn't necessarily mean that they actually are: volkit 
\emph{might}
resort to a more efficient implementation if the behavior stays unaffected.

\emph{Transform algorithms} (not to be confused with the \emph{algorithm}
\texttt{Transform}) are used to \texttt{Flip}, \texttt{Scale}, or
\texttt{Rotate} the volume with subcell accuracy.
\emph{Domain decomposition algorithms} decompose the volume into smaller parts. The
\texttt{BrickDecompose} algorithm for example decomposes
a given volume into a number of subgrids and optionally computes
ghost cells for interpolation. Volkit also includes statistical algorithms such as 
\texttt{ComputeHistogram} or
\texttt{ComputeAggregates}---the latter are, e.g., the min and max cell values and their
respective argmin and argmax, the average and standard deviation of the cell values,
etc. Arithmetic algorithms are used to compute sums, products, etc. on the
whole volume or on subranges.

\subsubsection{Rendering}
Volume rendering is supported through the
\texttt{Render} algorithm that will invoke an interactive, windowed ray tracer.
The algorithm can
be configured to perform either volume ray marching with absorption and 
emission, implicit iso-surface ray casting,
or volumetric path tracing with an isotropic
phase function.
We used the C++ library Visionaray~\cite{zellmann:2017b} for the implemenation
to target x86-64 and ARM CPUs as well as NVIDIA GPUs. Passing an RGBA lookup
table to the algorithm will invoke an interactive transfer function editor
in the same window as the ray tracer. The \texttt{RenderState} object parameter
is used to apply general settings (cf.~\cref{fig:render}).
In general, the algorithm is meant to be used for prototyping
to obtain high-quality renderings with little effort.
\begin{figure}[tb]
\begin{lstlisting}[language=c++]
/* --- ANSI C99 --------------------- */
vktStructuredVolume volC;
vktStructuredVolumeCreate(&volC,64,64,64,     /* dims */
                          vktDataFormatUint8, /* format */
                          1.f,1.f,1.f,        /* cell size */
                          0.f,1.f);           /* range */
vktFillRangeSV(volC,1,1,1,63,63,63,1.f); /* SV: structured */
vktStructuredVolumeDestroy(volC);

/* --- C++03 ------------------------ */
vkt::StructuredVolume volCPP(64,64,64,vkt::DataFormat::UInt8);
vkt::FillRange(volCPP,{1,1,1},{63,63,63},1.f);
\end{lstlisting}
\vspace{-1em}
\caption{\label{fig:exampleCvsCPP} ANSI C99 and C++03 APIs, the code
snippets demonstrate how to use the two APIs, and the major differences
when performing equivalent computations.}
\end{figure}
\subsection{Language Bindings}
Volkit is primarily a C/C++ library, but comes with bindings for other
languages. Internally, volkit is implemented in  C++14 and
optionally uses language extensions such as NVIDIA CUDA.
We however ship header files with separate interfaces for 
C99 and for C++03. The two interfaces are designed to expose the same 
functionality, but the C++03 interface adds a couple of convenience functions
such as RAII for managed types, function overloads for short vectors, etc.
Exemplary use of the two interfaces is shown in
\cref{fig:exampleCvsCPP}.

We also provide Python~3 bindings that were realized using SWIG~\cite{beazley:1996};
in theory, that should also allow us to support other languages but
so far hasn't been tested. Volkit has a command
line interface where algorithms are invoked via separate
processes. The base command is called \texttt{vkt}. In order for
example to render a volume, a volume file is \emph{piped} into the process,
which is called with the 
argument \texttt{render}. Inter-process communication is
implemented with file I/O. The user can---and is encouraged 
to---use shared memory to realize that communication to
bypass the file system and costly hard drive accesses.

\begin{table}
\caption{Structured and hierarchical volume data sets we use for the evaluation.
See the papers by Lee et al.~\cite{lee:2013,lee:2015} for the DNS data set and
by Seifried et al.~\cite{seifried:2017} for the SILCC moledular cloud data set.
}\label{tab:eval}
\begin{tabular}{|l|l|}
\hline
\multicolumn{2}{|l|}{\textbf{Heptane (Gas)} type: structured}\\
\hline
\begin{tabular}{l}
Dims: 302$\times$302$\times$302\\Cell type: \texttt{UInt8}\\Total size: 26.3~MB\\
\end{tabular} & \parbox[c]{.7\textwidth}{\includegraphics[width=.7\textwidth]{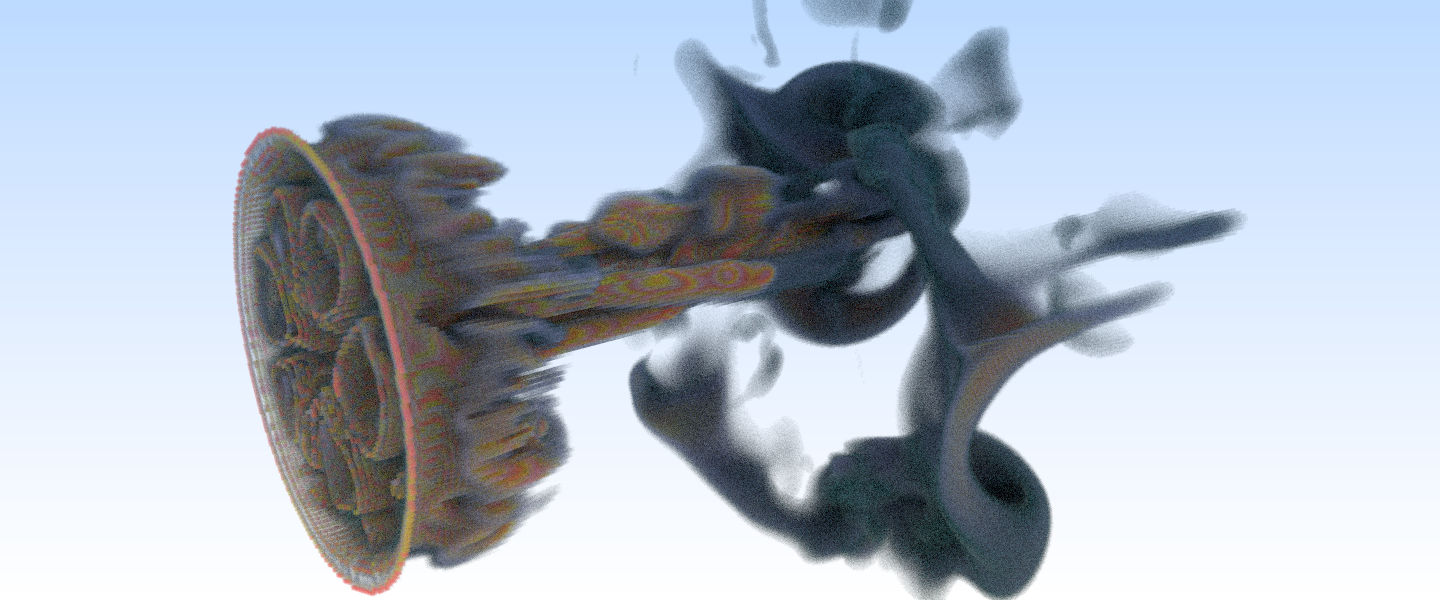}}\\
\hline
\multicolumn{2}{|l|}{\textbf{Turbulent Channel Flow Sim (DNS)} type: structured}\\
\hline
\begin{tabular}{l}
\textbf{DNS~4}\\Dims: 2560$\times$1920$\times$384\\Cell type: \texttt{Float32}\\Total size: 7.1~GB\\
\hline
\textbf{DNS~8}\\Dims: 1280$\times$960$\times$192\\cell type: \texttt{Float32}\\Total size: 900~MB\\
\end{tabular} & \parbox[c]{.7\textwidth}{\includegraphics[width=.7\textwidth]{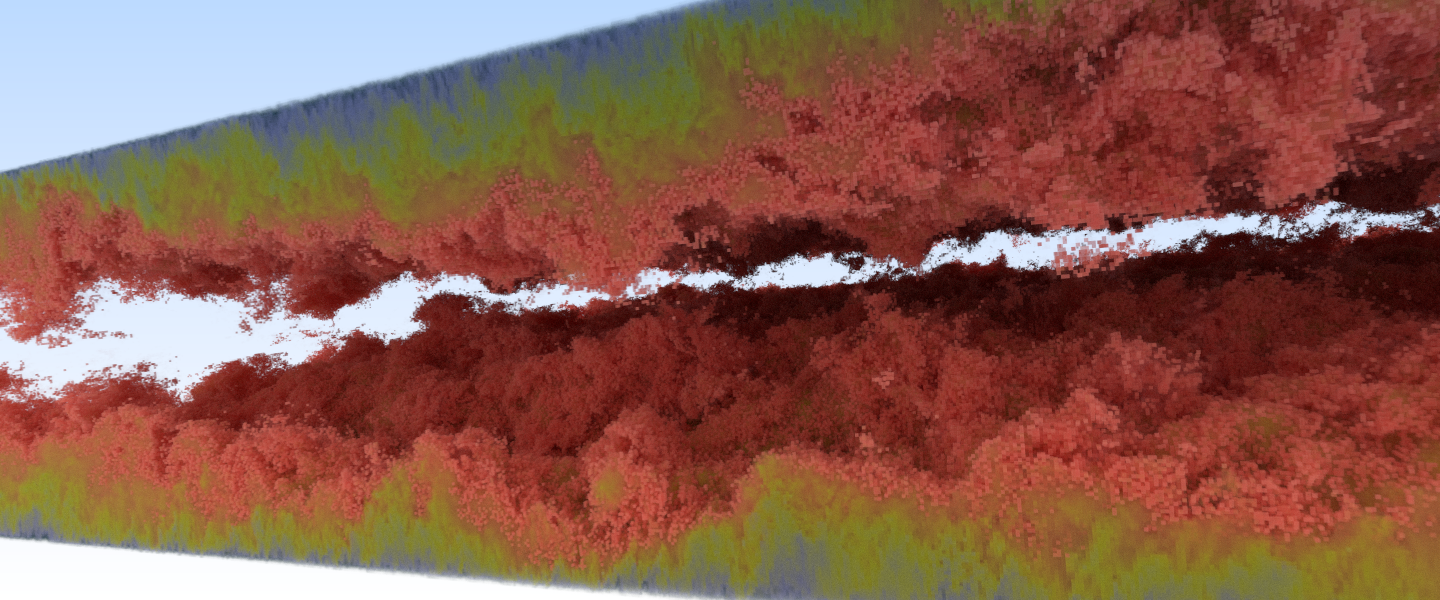}}\\
\hline
\multicolumn{2}{|l|}{\textbf{Richtmyer-Meshkov Instability (RM)} type: structured}\\
\hline
\begin{tabular}{l}
Dims: 2048$\times$2048$\times$1920\\Cell type: \texttt{UInt8}\\Total size: 7.5~GB\\
\end{tabular} & \parbox[c]{.7\textwidth}{\includegraphics[width=.7\textwidth]{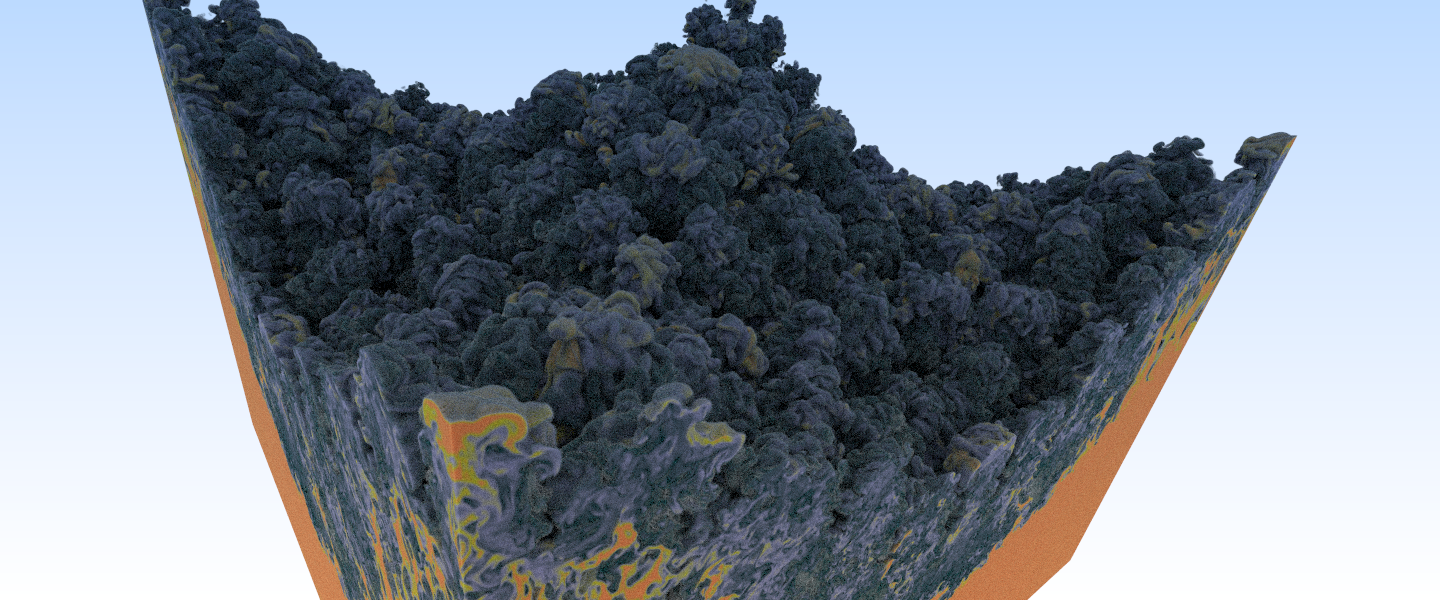}}\\
\hline
\hline
\multicolumn{2}{|l|}{\textbf{SILCC Molecular Cloud (ARM)} type: hierarchical}\\
\hline
\begin{tabular}{l}
Logical grid dims:\\
4096$\times$4096$\times$81920\\Cell type: \texttt{Float32}\\
Cells: 72.8M\\
Subgrids: 142K\\
Levels: 7\\
Total size: 283~MB\\
\end{tabular} & \parbox[c]{.7\textwidth}{\includegraphics[width=.7\textwidth]{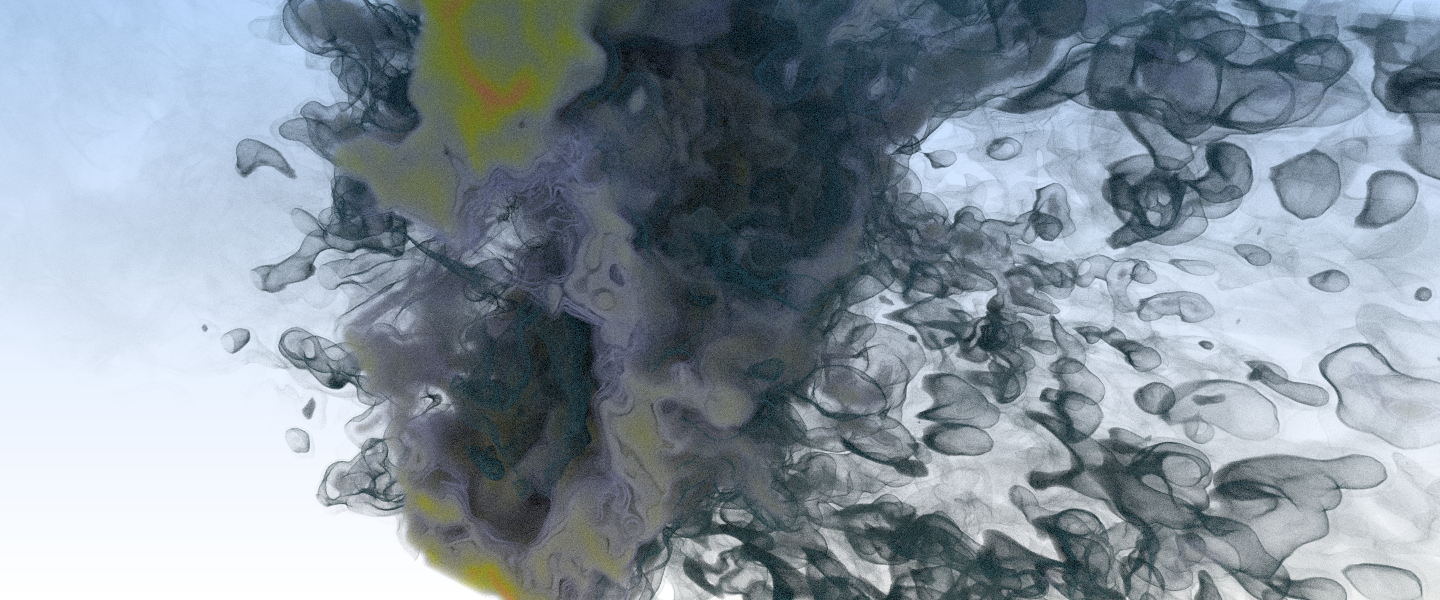}}\\
\hline
\end{tabular}
\end{table}

\section{Evaluation and Case Studies}
\begin{figure}[t]
\centering
\includegraphics[width=0.48\textwidth]{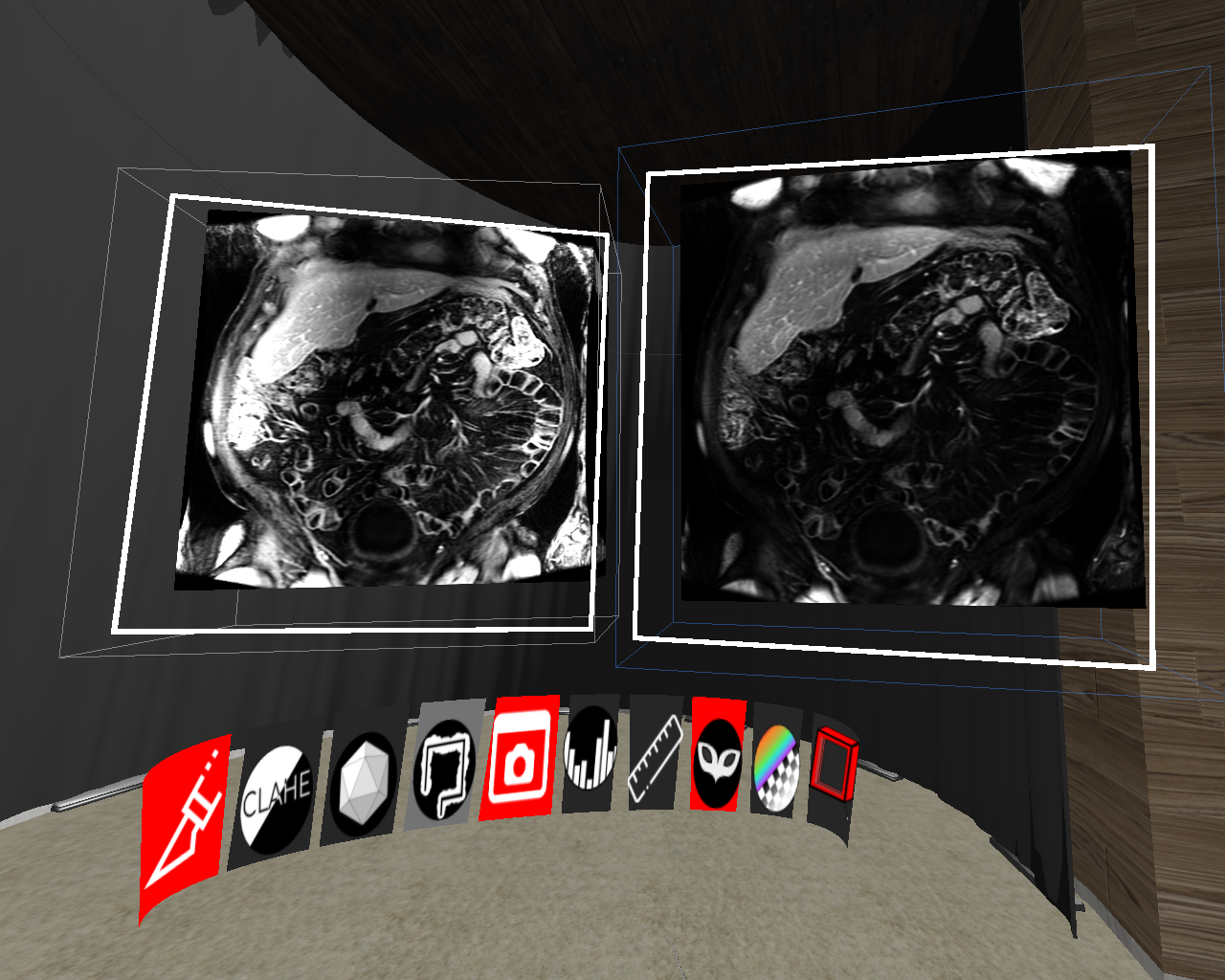}~~
\includegraphics[width=0.48\textwidth]{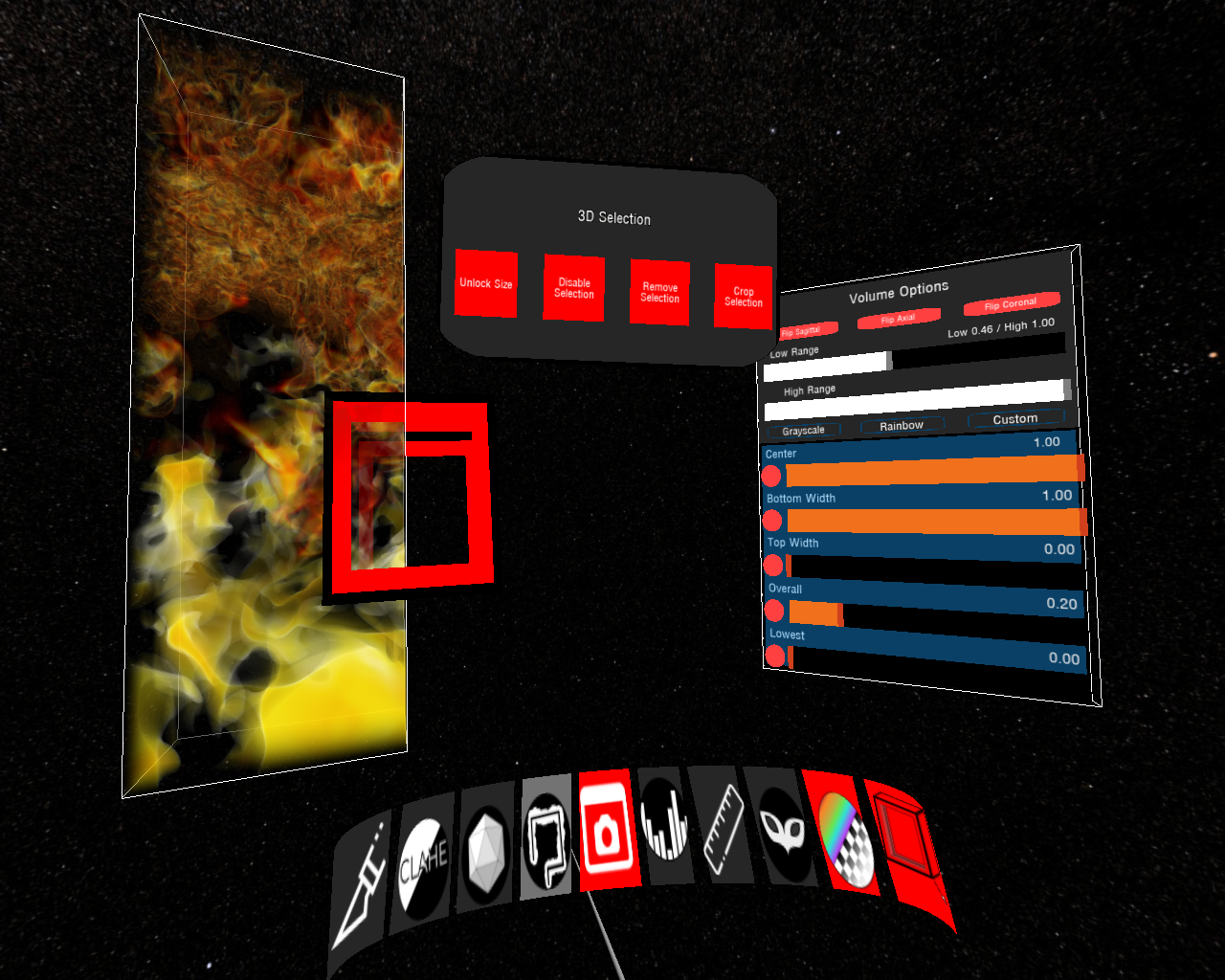}
\vspace{-1em}
\caption{\label{fig:casestudies}
Volkit / CalVR virtual reality case studies. Left: side-by-side view of contrast-enhanced medical image rendering using CLAHE (left volume) and unaltered original (right volume)
Right: AMR volume exploration with a region of interest (ROI).
Inside the ROI, the volume is resampled on a structured grid with high resolution.}
\end{figure}
We present two case studies
as well as a performance study to assess volkit's effectiveness and efficiency.
For the latter we tested an assortment of algorithms using multiple real-world
data sets (cf.~\cref{tab:eval}).

\subsection{Case Study: CLAHE-3D}
For this case study we implemented a 3D variant of the Contrast Limited 
Adaptive Histogram Equalization (CLAHE)
algorithm~\cite{pizer:1987} found in OpenCV~\cite{bradski:2013} using
volkit's CPU and CUDA backends. CLAHE is used for contrast
enhancement and applies histogram equalization to subregions of the volume.
This is done via a brick decomposition and local
histograms. The ensuing histogram equalization step uses a maximum cut-off
to reduce noise in the resulting volume---noise
reduction is the major difference compared to other contrast enhancement
techniques. The local results are then combined using linear interpolation.

Those steps are implemented using individual CUDA kernels or,
on the CPU, loops over either the whole
volume, the bricks, or the local histograms. For the 
implementation we use the various routines provided by volkit that allow for
accessing and manipulating volumes. A contrast-enhanced medical 3D
image is shown in~\cref{fig:casestudies}. The VR software
CalVR~\cite{schulze:2013} that we integrated volkit into comes with
a GLSL implementation of CLAHE~\cite{lucknavalai:2020}. We found our CUDA version
to be slightly faster than CalVR's implementation (the
difference in performance we attribute to implementation details
regarding how CalVR executes compute shaders via scene graph nodes),
where the performance for volumes of $256^3$ cells would be on the order of 30~ms.
The CPU version we found to be about one to two orders of magnitude
slower than the CUDA implementation.

\subsection{Case Study: Exploring Large Simulation Data with CalVR}
Our second case study is comprised of integrating volkit into the virtual reality
software CalVR~\cite{schulze:2013}, which is targeted towards large back projection
and tiled display systems, but also towards VR headsets.
CalVR comes with a volume rendering plugin
that we extended with a zoom interaction for AMR volume exploration.
To achieve frame rates high enough for VR, we don't render
the AMR volume directly, but instead resample it on a structured
grid in an interactive preprocess. Initially, we resample the AMR volume to
a structured volume based on a user-defined target number of cells---for our
experiments we used 50~M. The user then specifies a region of interest (ROI)
to zoom in on (cf.~\cref{fig:casestudies}). The volume is cropped accordingly
using the \texttt{Crop} algorithm, giving us a smaller
AMR volume, which we resample using \texttt{Resample}
and the same number of target cells from before. This effectively allows us to zoom in on the 
volume and by that to increase the resolution inside the ROI.
We experimented with the data set from~\cite{seifried:2017}
depicted in \cref{fig:silcc,fig:casestudies,fig:eval}, whose logical grid 
size is 4K$\times$4K$\times$80K
and found the zoom interaction to always take on the order of below
one second.

\subsection{Performance}
\label{sec:eval}
\begin{figure}
\centering
\includegraphics[width=.3\textwidth]{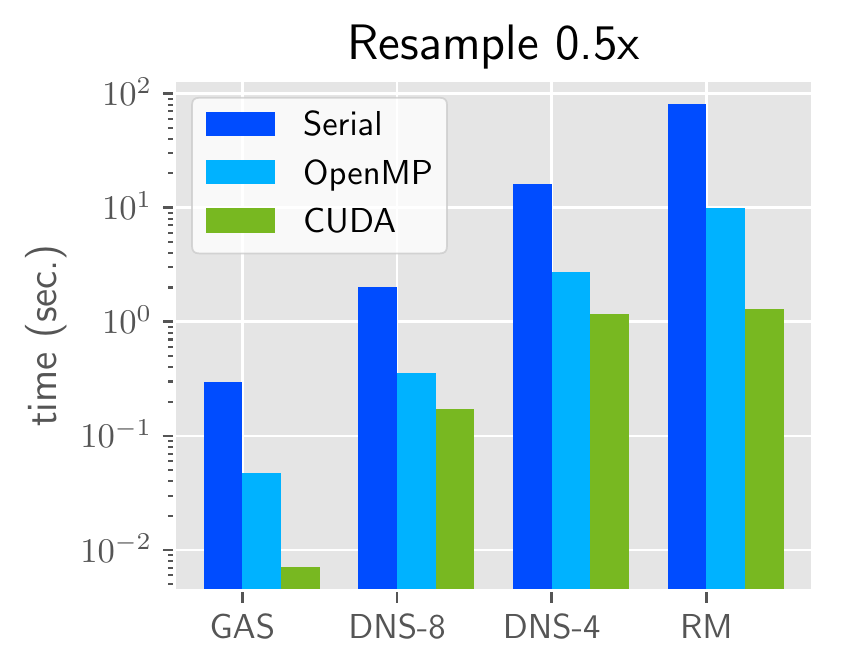}
\includegraphics[width=.3\textwidth]{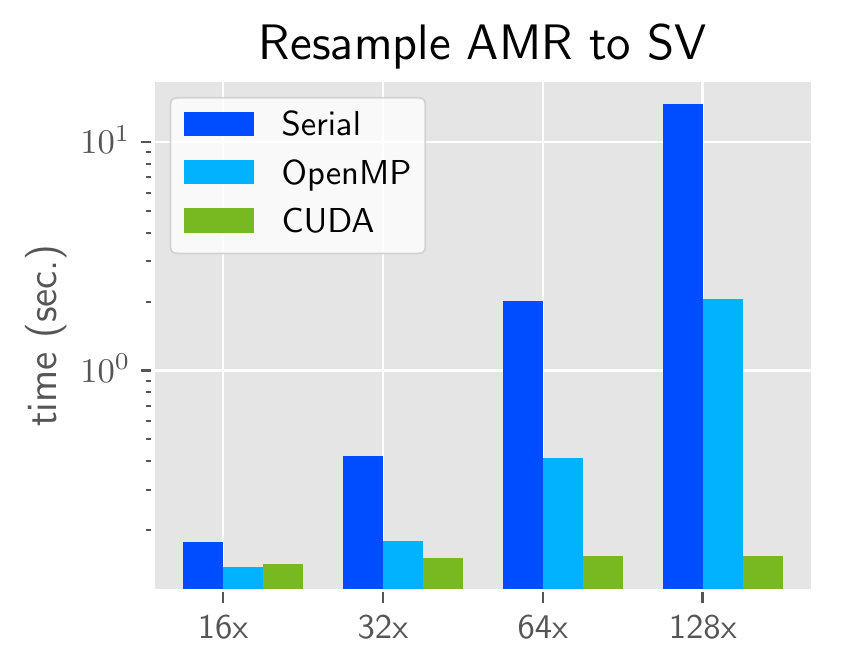}
\includegraphics[width=.3\textwidth]{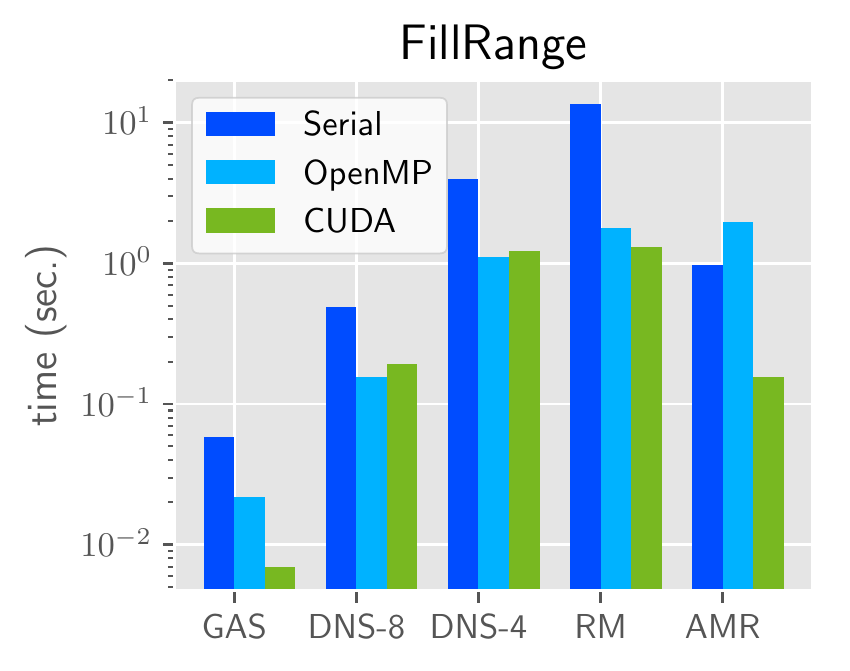}\\
\includegraphics[width=.3\textwidth]{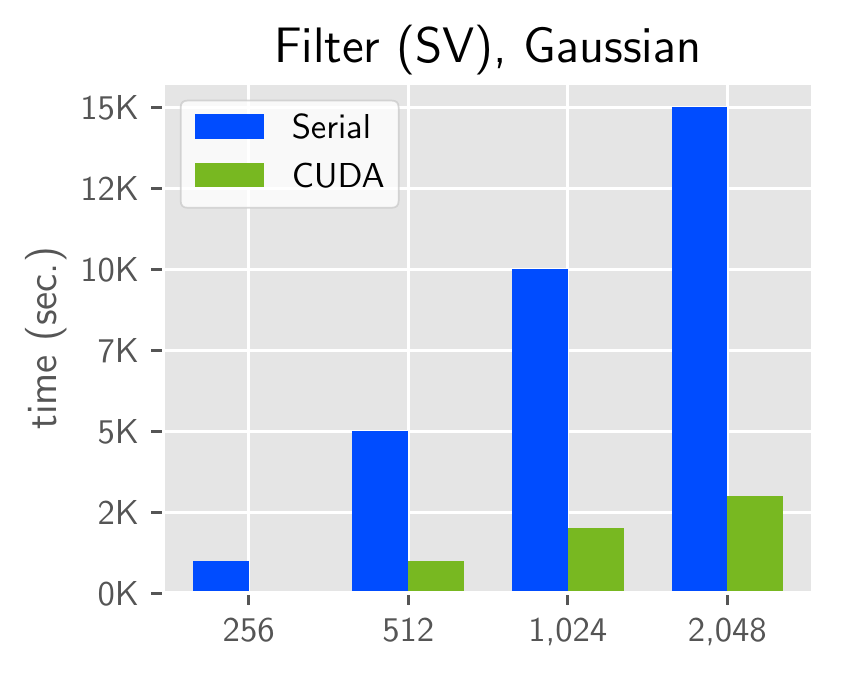}
\includegraphics[width=.3\textwidth]{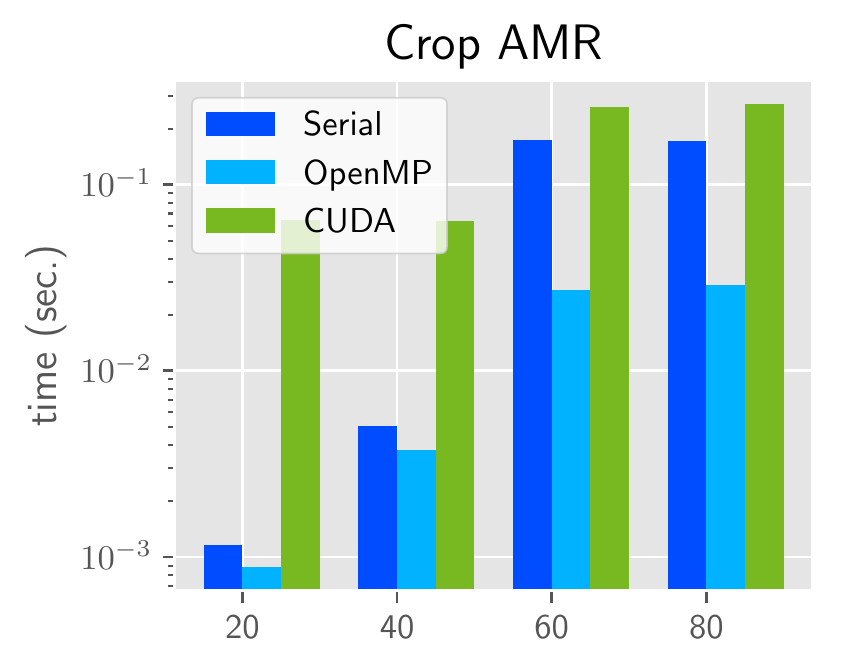}
\includegraphics[width=.3\textwidth]{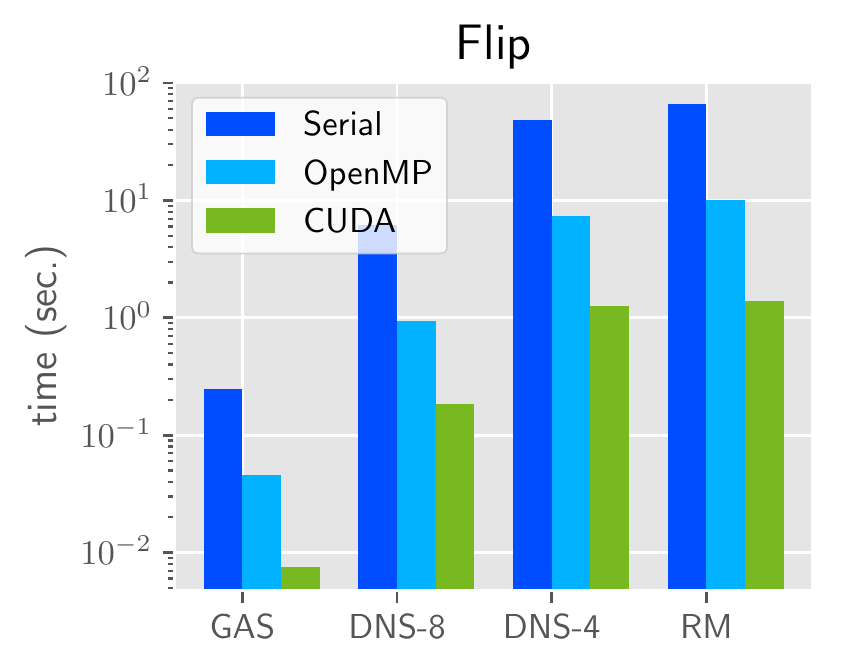}
\vspace{-1em}
\caption{Single-threaded, multi-threaded (16x, with OpenMP) and GPU results for an 
assortment
of algorithms implemented in volkit. Note the logarithmical scale to make
the scalability of the CPU and GPU results comparable.}
\vspace{-1em}
\label{fig:eval}
\end{figure}
We present performance numbers for some of the algorithms implemented in volkit.
Our main motivation here is not to evaluate the individual algorithms---we chose an
assortment of algorithms that, due to their characteristics will scale more or less well---but
rather to prove that our design decisions---deferred API, data management layer, choice of
programming environment, etc., don't severely impact overall performance and scalability.

We use \texttt{Resample} to scale
down the structured volumes shown in~\cref{tab:eval} by a factor of two in each dimension; \texttt{Resample} to
convert an AMR volume (also~\cref{tab:eval}) to structured volumes of different sizes; \texttt{FillRange} to
fill the whole volume with a uniform value;
\texttt{ApplyFilter} to apply Gaussian blur to the structured volumes; \texttt{Crop}
with with a sliding window and ROIs of 20\%, 40\%, 60\%, and 80\% of the original
AMR volume's size; and \texttt{Flip} to
mirror the structured volumes along the axis where their extents are longest.
We ran single-threaded and multi-threaded tests on a 16 core Intel Xeon
CPU with 2.2~GHz that is equipped with 128~GB DDR memory and an NVIDIA Quadro RTX~8000 GPU
with 12~GB of GDDR memory. Results are presented in~\cref{fig:eval}.
We observe that the GPU outperforms the CPU for structured volumes
due to superior memory bandwidth. For AMR volumes we see mixed results---this
is likely due to the fact that most of our algorithms parallelize over subgrids,
so that the problem is less balanced and exposes less parallelism.

\section{Conclusions}
We presented the performance-portable 3D volume manipulation library volkit
targeted at large volumes that are for example generated
by simulation codes. In contrast to other libraries, volkit focuses on particular volume representations
such as for example hierarchical / AMR data. We presented volkit's overall system design,
as well as case studies and evaluations. Finally,
volkit is open source and can be downloaded from~\url{https://github.com/volkit/volkit}.

\section*{Acknowledgments}
\iffinal
This research was funded in part by a grant from The Leona M. and Harry B. Helmsley Charitable 
Trust. The SILCC AMR data set is courtesy Daniel Seifried. The structured volume data sets are
hosted online by Pavol Klacansky under~\url{https://klacansky.com/open-scivis-datasets/}.
\else
Intentionally omitted for review.
\fi
%
%

\bibliographystyle{splncs04}
\bibliography{stefan}

\end{document}